\documentclass[conference]{IEEEtran}
\IEEEoverridecommandlockouts

\usepackage{amsmath,amssymb,amsfonts}
\usepackage{algorithmic}
\usepackage{graphicx}
\usepackage{tikz}
\usepackage{textcomp}
\usepackage{cite}

\makeatletter
\def\@citex[#1]#2{\leavevmode
\let\@citea\@empty
\@cite{\@for\@citeb:=#2\do
{\@citea\def\@citea{,\penalty\@m\ }%
\edef\@citeb{\expandafter\@firstofone\@citeb\@empty}%
\if@filesw\immediate\write\@auxout{\string\citation{\@citeb}}\fi
\@ifundefined{b@\@citeb}{\hbox{\reset@font\bfseries ?}%
\G@refundefinedtrue
\@latex@warning
{Citation `\@citeb' on page \thepage \space undefined}}%
{\@cite@ofmt{\csname b@\@citeb\endcsname}}}}{#1}}
\makeatother
    
\begin{document}

\newcommand\copyrighttext{%
  \footnotesize This work has been submitted to the IEEE for possible publication. Copyright may be transferred without notice, after which this version may no longer be accessible.}%
\newcommand\copyrightnotice{%
\begin{tikzpicture}[remember picture,overlay]%
\node[anchor=south,yshift=10pt] at (current page.south) {\fbox{\parbox{\dimexpr\textwidth-2cm}{\copyrighttext}}};%
\end{tikzpicture}%
\vspace{-10pt}%
}

\title{Robust Semantic Segmentation in Adverse Weather Conditions by means of Sensor Data Fusion}
\author{\IEEEauthorblockN{ Andreas Pfeuffer, and Klaus Dietmayer}
\IEEEauthorblockA{Institute of Measurement, Control, and Microtechnology, Ulm University, 89081 Ulm, Germany} 
Email: \{andreas.pfeuffer, klaus.dietmayer\}@uni-ulm.de}

\newcommand\Tstrut{\rule{0pt}{2.0ex}}         
\newcommand\Bstrut{\rule[-0.9ex]{0pt}{0pt}} 
\newcommand\Mstrut{\rule[-0.0ex]{0pt}{0pt}}

\maketitle
\copyrightnotice

\begin{abstract}
	A robust and reliable semantic segmentation in adverse weather conditions is very important for autonomous cars, but most state-of-the-art approaches only achieve high accuracy rates in optimal weather conditions. The reason is that they are only optimized for good weather conditions and given noise models. However, most of them fail, if data with unknown disturbances occur, and their performance decrease enormously. 
	One possibility to still obtain reliable results is to observe the environment with different sensor types, such as camera and lidar, and to fuse the sensor data by means of neural networks, since different sensors behave differently in diverse weather conditions. Hence, the sensors can complement each other by means of an appropriate sensor data fusion. Nevertheless, the fusion-based approaches are still susceptible to disturbances and fail to classify disturbed image areas correctly. This problem can be solved by means of a special training method, the so called Robust Learning Method (RLM), a method by which the neural network learns to handle unknown noise.
	In this work, two different sensor fusion architectures for semantic segmentation are compared and evaluated on several datasets. Furthermore, it is shown that the RLM increases the robustness in adverse weather conditions enormously, and achieve good results although no disturbance model has been learned by the neural network.  
\end{abstract}


\section{Introduction}

\begin{figure}[tbp]
		\includegraphics[width=1.0\columnwidth]{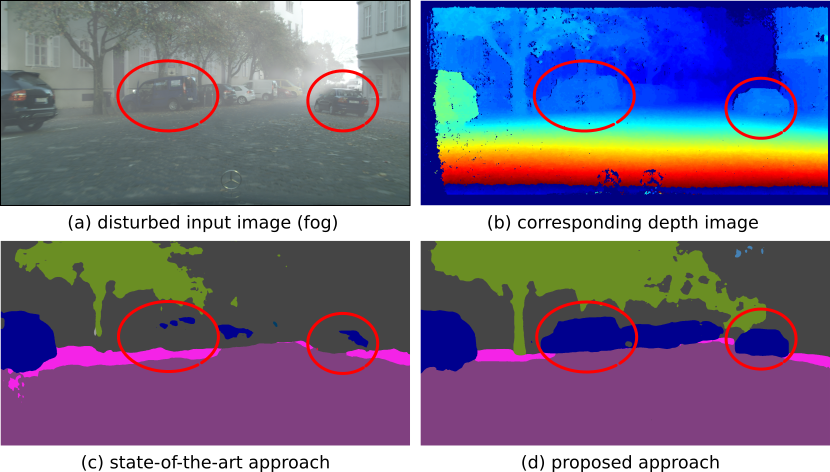}
		\caption{Qualitative comparison of the state-of-the-art training method with proposed learning technique by means of the Foggy Cityscapes Dataset.}
		\label{fig_result_introduction}
		\vspace{-5mm}
	\end{figure}

	A reliable environment recognition in adverse weather conditions is a quite challenging task for autonomous driving. While many applications, such as the semantic segmentation of the camera streams, achieve very good results in good weather conditions, they fail in non optimal weather conditions when the sensors are disturbed by rain, snow, fog or even by dazzling sun, and their accuracy decreases enormously.
	Usually, the sensors of autonomous cars fail asymmetrically in bad weather conditions. For instance, in blinding sunlight, the camera images are disturbed by large white areas so that there is no information available about the environment. In contrast, the corresponding environment can still be observed by the lidar sensor, which is not disrupted by blinding sun.
	This observation motivates us to use a multiple sensor-setup, e.g camera image and lidar data, or the image of a stereo camera and the corresponding depth image, to increase the robustness of the semantic labeling approaches.
	However, the robustness of these approaches is not necessarily improved by the fusion of multiple sensor data, as the example in Fig. \ref{fig_result_introduction} shows. In this example, the camera image is artificially disturbed by fog so that the cars in the image are less visible, while the corresponding depth image is not disturbed. It is expected that the cars are still recognized completely, since the cars can be clearly seen in the depth image. However, large parts of the cars are not classified correctly as car but as building and road. 
	The reason is that state-of-the-art approaches are particularly vulnerable if one sensor stream is corrupted by noise the neural network has not seen during training. By simulating a special noise pattern, the performance can be increased for this type of noise. For example, Sakaridis et al. \cite{Sakaridis_2018_SemanticFoggySceneUnderstandingWithSyntheticData} have created synthetic fog data, and have shown that they achieve a greater accuracy on real foggy data.
	However, it is almost impossible to record or simulate training data covering all adverse weather scenarios and all possible sensor disturbances, since the noise in diverse weather conditions is rather complex. 
	Therefore, a suitable training method is necessary so that the network learns to focus on the undisturbed sensor data and ignores the disrupted sensor data of unknown noise types.

	In this work, a robust semantic segmentation approach is presented, which can deal with different adverse weather situations and unknown noise pattern due to an advanced training method, the so called Robust Learning Method (RLM). The proposed approach is evaluated on different datasets and by means of various noise types, which have not been used for training. It turns out that the proposed approach outperforms other state-of-the-art approaches. For instance, the proposed approach can classify the cars in the introductory example (see Fig. \ref{fig_result_introduction}) completely, while the state-of-the-art approach only classifies a tiny fractions of the cars correctly. Moreover, two different fusion architectures (Early and Late Fusion) are investigated to find out which fusion strategy is more appropriate for a robust semantic segmentation.


\section{Related Work}
\label{chap_RelatedWork}

	In recent years, convolution neural networks (CNNs) are widely used for computer vision tasks such as semantic segmentation approaches, which assign  every pixel of the image a predefined class. Early CNN-based approaches are originally derived from the image classification task and often consist of an encoder-decoder architecture \cite{2015_Badrinarayanan_SegNet_ADeepConvolutionalEncoderDecoderARchitectureForImageSegmenatation, Long_2015_FullyConvolutionalNetworksForSemanticSegmentation}. 
	Their performance has continuously improved in several ways, e.g. by multi-scale architectures \cite{Chen_2017_DeepLab_SemanticImageSegmentationWithDeepConvolutionalNets_AtrousConvolution_andFullyConnectedCRFs, Chen_2018_Deeplabv3p_EncoderDecoderWithAtrousSeparableConvolutionForSemanticImageSegmentation, Zhao_2017_PyramidScenParsingNetwork}, or the usage of temporal information \cite{Valipour_2017_RecurrentFullyConvolutionalNetworksForVideoSegmentation, Yurdakul_2017_SemanticSegmentationOfRGBDVideosWithRecurrentFullyConvolutionalNeuralNetworks}. 
	Additionally, some approaches focus on the reduction of the inference time \cite{Paszke_2016_ENet_ADeepNeuralNetworkArchitectureForRealTimeSemanticSegmentation, Zhao_2017_ICNet_forRealTimeSemanticSegmentationOnHighResolutionImages}, so that they are practicable for real-time applications such as autonomous driving or mobile robots.
	Despite pure camera-based semantic segmentation approaches, there are also semantic segmentation approaches for other sensors, e.g. for lidar \cite{Caltagirone_2017_FastLIDAR-basedRoadDetectionUsingFullyConvolutionalNeuralNetworks, Dewan_2017_DeepSemanticClassificationFor3DLiDARData, Piewak_2018_BoostingLiDAR-basedSemanticLabelingByCross-ModalTrainingDataGeneration}, or for radar \cite{Schumann_2018_SemanticSegmentationOnRadarPointClouds}.
	
	The segmentation accuracy can be increased further by using information of several sensors. For instance, the fusion of RGB and depth image of a stereo camera is very popular in the literature. 
	Generally, there are two possibilities to fuse image and depth information by means of neural networks. On the one hand, 2D semantic image segmentation approaches are extended by an additional channel for the depth image \cite{Ma_2017_Multi-ViewDeepLearningForConsistentSemanticMappingWthRGB-DCameras, Wang_2016_LearningCommonAndSpecificFeaturesFoRGB-DSemanticSegmentationWithDeconvolutionalNetworks}, which is also known as early fusion in the literature. On the other hand, the fusion is performed at feature level, which is denoted as late fusion. For example, FuseNet \cite{Hazirbas_2016FuseNet_IncorporatingDepthIntoSemanticSegmentationViaFusion-BasedCNNArchitecture} determines two separate feature maps for camera and depth image by means of two VGG16-encoders, which are combined before they are fed into a common decoder. 
	In \cite{Lee_2017_RDFNetRGB-DMulti-levelResidualFeatureFusionForIndoorSemanticSegmentation}, camera and depth image are processed by two independent neural networks, and the two resulting predictions are combined in the end. 
	Although fusing camera and lidar data is very popular in object detection tasks \cite{Chen_2016_MultiView3DObjectDetectionNetworkForAutonomousDriving, Ku_2018_Joint3DProposalGeneartionAndObjectDetectionFromViewAggregation, 2018_Pfeuffer_OptimalSensorDataFusionArchitectureForObjectDetectionInAdverseWeatherConditions}, there are hardly any similar approaches for semantic segmentation, which is mainly caused by the fact that a public available dataset for semantic segmentation containing camera and lidar data does not yet exist. Nevertheless, a fusion approach for semantic labeling is introduced in this work, which is based on camera and lidar data, and is evaluated on our in-house dataset.
	
	Most semantic segmentation and computer vision approaches focus on achieving high scores at well-known benchmarks, but they do not care much about the robustness of their methods in adverse weather conditions. In fact, there are only a few approaches which can deal with disturbances caused by diverse weather often focusing on one special noise type. 
	For instance, Sakaridis et al. \cite{Sakaridis_2018_SemanticFoggySceneUnderstandingWithSyntheticData} increase the segmentation accuracy in fog by adding simulated fog to real-world images, and by training with these synthetic data. Porav et al. \cite{Porav_2019_ICanSeeClearlyNow_ImageRestorationViaDeRaining} derain rainy images by means of a neural network and achieve better performance, when a further segmentation network is applied on the derained images. 
	Bijelic et al. \cite{Bijelic_2018_RobustnessAgainstUnknownNoiseForRawDataFusingNeuralNetworks} introduces the Robust Learning Technique, by which an image classifier becomes more robust against unknown disturbances not occurring in the training set. The authors randomly chose one of the input channels (camera image of depth image) and replace the corresponding input data by an arbitrarily selected sensor data of another, different scene. 
	In \cite{2018_Pfeuffer_OptimalSensorDataFusionArchitectureForObjectDetectionInAdverseWeatherConditions}, a training strategy is described which increases the robustness of an object detector in diverse weather conditions by fitting white polygons into the training data. 
	Based on this training strategy, a robust semantic segmentation approach is proposed hereinafter, which achieves good results on disturbed adverse-weather data.


\section{Network Architectures}

	In this section, the proposed sensor fusion architectures are described, which are based on the ICNet \cite{Zhao_2017_ICNet_forRealTimeSemanticSegmentationOnHighResolutionImages}, a pure camera-based semantic segmentation approach.  
	The ICNet is a real-time capable semantic segmentation method, which still delivers high performance by progressively processing the input image in multiple resolutions. More precisely, the input image is downsampled twice so that three different branches with input images of scale $1$, $1/2$ and $1/4$ are yielded. Each branch is processed independently and combined at the end of the network by means of Cascade Feature Fusion (CFF) layers \cite{Zhao_2017_ICNet_forRealTimeSemanticSegmentationOnHighResolutionImages}. First, the determined feature maps of the low resolution branch and the medium resolution branch are fused using a CFF layer. Then, the output of the CFF layer is concatenated with the feature map of the high resolution branch by means of an additional CFF layer. The fused resolution branches are upsampled to the size of the input image using interpolation and further convolution layers. Finally, a softmax classifier is applied to predict the class of each image pixel.
	The described origin ICNet is extended so that different sensor data, e.g. camera and lider, can be fused by means of the neural network. First of all, the camera image and the lidar data have to be preprocessed so that they are in an appropriate input format.
	The camera image is first resized to an appropriate size and the image mean is subtracted from the image, so that it is independent from the image illumination. 
	The lidar data is used to determine a dense depth image similar to the disparity map of a stereo camera, as described in \cite{2018_Pfeuffer_OptimalSensorDataFusionArchitectureForObjectDetectionInAdverseWeatherConditions}.
	For this, the lidar data is first filtered so that only those points located in the camera field of view are left. The remaining 3D points are then projected onto the image plane so that a sparse depth map results. In a next step, the pixels without depth information are filled by interpolating between neighboring points. Note, if there is no point within a predefined neighborhood, the corresponding pixel is set to infinity.  
	In this work, two different network architectures for an appropriate sensor fusion are considered, the so called Early Fusion and the Late Fusion. 
	In the Early Fusion, the preprocessed sensor data are fused right at the beginning, and in the Late Fusion, the sensor data are fused at the end of the neural network. Both fusion strategies are described in more detail in the next sections.

	\begin{figure*}[tbp]
		\includegraphics[width=1.0\textwidth]{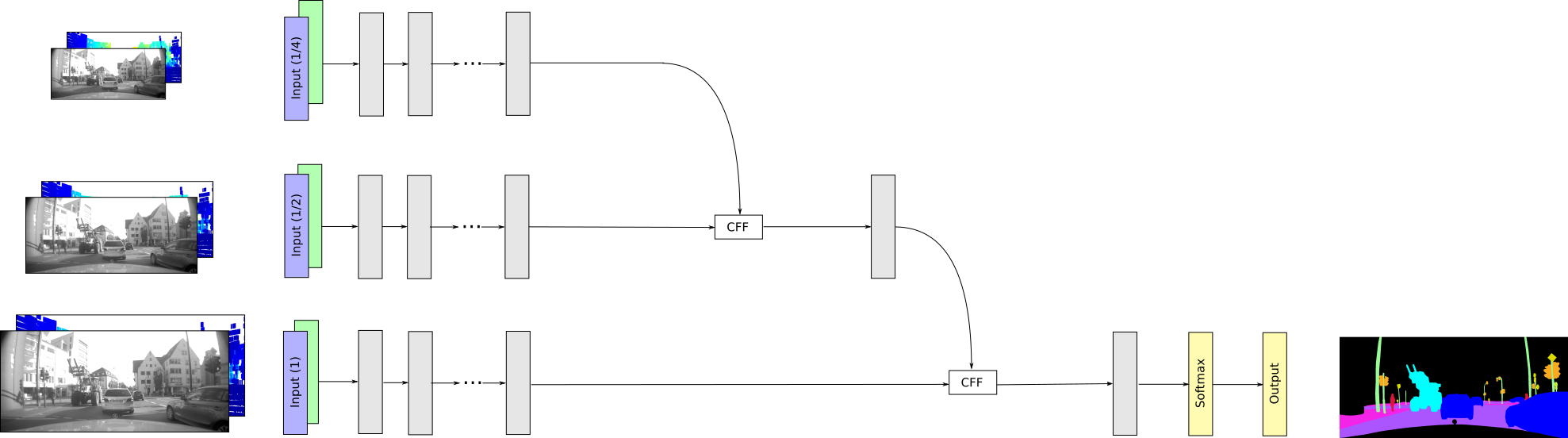}
		\caption{Early Fusion Approach: RGB image and the depth image determined from the lidar data are fit into a common 4D tensor and progressed together in the entire neural network}
		\label{fig_architecture_EarlyFusion}
	\end{figure*}
	
	\begin{figure*}[tbp]
		\includegraphics[width=1.0\textwidth]{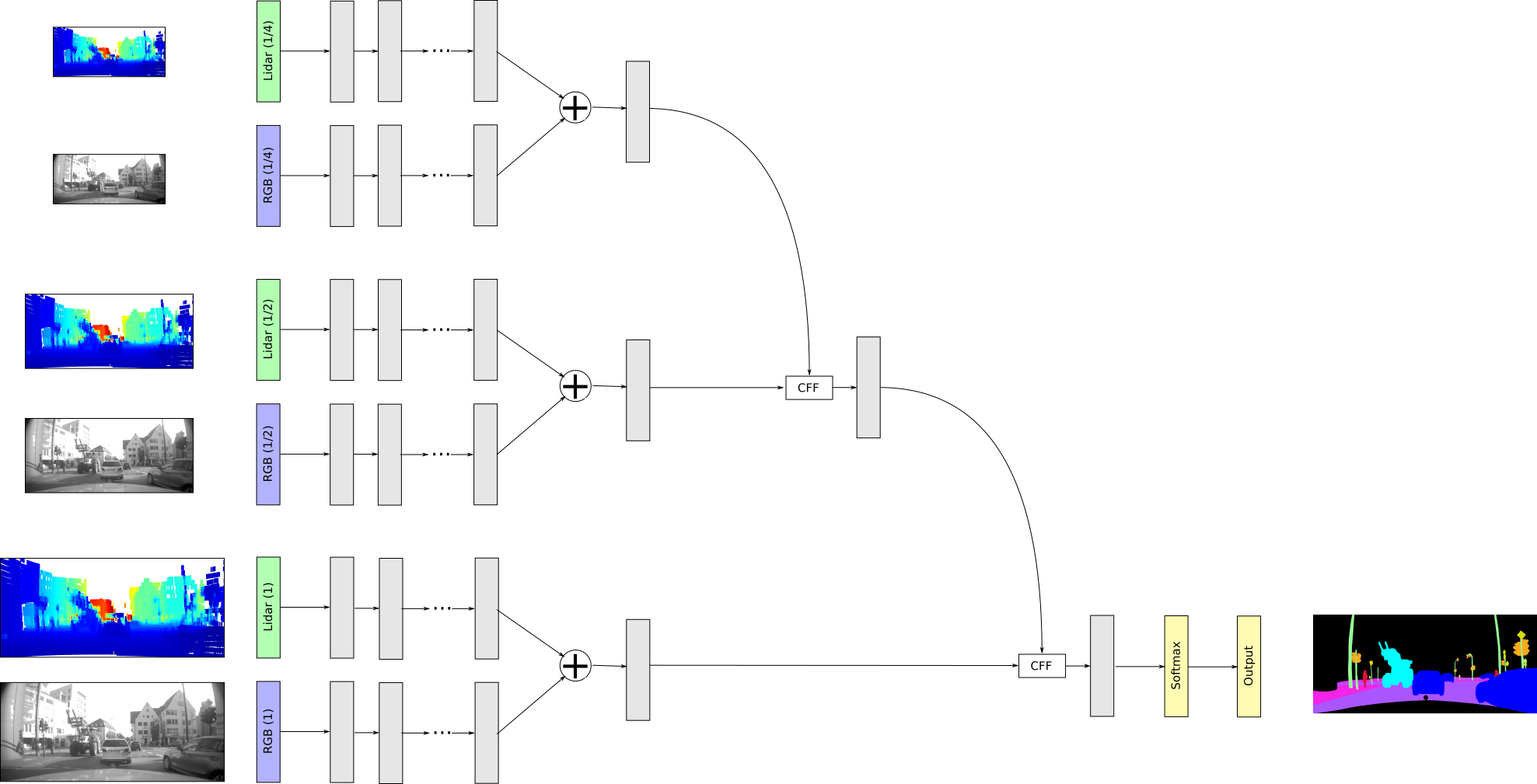}
		\caption{Late Fusion Approach: For each sensor and for each resolution, an independent feature map is determined by distinct feature encoders. The camera and lidar feature maps of one scale are then concatenated before the different scales are fused by means of CFF layers.}
		\label{fig_architecture_LateFusion}
	\end{figure*}

\subsection{Early Fusion}
	
	The idea of the Early Fusion is, that the camera image and the corresponding depth image are concatenated to a common input tensor and are processed together throughout the network, so that the neural network can learn by itself how to fuse the sensor data. For this, the input dimension of the origin ICNet is extended from 3D to 4D. The first three dimensions contain the three color channels of the camera image and the 4th dimension contains the determined dense depth image. The common 4D input tensor is fed into the three branches with different scales to yield the semantic labels of the corresponding input data. The net architecture of the Early Fusion is illustrated in Fig. \ref{fig_architecture_EarlyFusion}.

\subsection{Late Fusion}

	Another fusion strategy is the Late Fusion, where the different sensor data streams are first processed separately before they are fused at the end of the neural network. The motivation is that each sensor has sensor specific properties, which might get lost in the Early Fusion. If each sensor is handled independently instead, the senor specific properties can be considered better.
	Another advantage is that disturbances of one sensor do not affect the other sensors until the fusion of their feature maps. 
	Generally, the architecture of the Late Fusion is similar to the original ICNet, but the amount of resolution branches is doubled, where two branches have got the same resolution $s$ ($s \in \{1; 1/2; 1/4\}$). The camera image is fed into one branch of resolution $s$, while the preprocessed lidar data is fed into the other branch of resolution $s$. The feature maps of camera and lidar data of resolution $s$ are concatenated, before they are fused with the other feature maps of different resolutions by means of the CFF layers. The described Late Fusion architecture is shown in Fig \ref{fig_architecture_LateFusion}.


\section{Robust Learning Method}

	The problem of most state-of-the-art approaches fusing different sensor data by means of neural networks is that they are quite vulnerable to sensor disturbances, which do not occur during training, and hence, the performance of these approaches usually decreases in these scenarios enormously.	
	An intuitive explanation is that the neural network has learned a joint representation of the individual sensor streams, e.g. if the object can be clearly seen in all sensor data, it can be classified correctly. However, if one sensor stream is disturbed by unknown noise, the object cannot be classified correctly. Therefore, the goal is that the neural network should learn an ``OR''-connection instead of an ``AND''-connection. In other words, the neural network should still function correctly, if one sensor is disrupted by neglecting the disturbed data and using the undisturbed data instead. 
	Therefore, the training strategy described in \cite{2018_Pfeuffer_OptimalSensorDataFusionArchitectureForObjectDetectionInAdverseWeatherConditions} is used, which overcomes this problem. For this, the clear, undisturbed data recorded in optimal weather conditions are modified for training to increase the robustness of the neural network. Analogous to \cite{2018_Pfeuffer_OptimalSensorDataFusionArchitectureForObjectDetectionInAdverseWeatherConditions}, one of the sensor streams is randomly chosen and set to infinity to simulate a sensor failure. Furthermore, white polygons are randomly fitted in the camera or the depth image, which represents partially disturbed data. In total, the good weather data, partly disturbed data and completely impaired data are used at the ratio of 1:2:4 for training the proposed semantic segmentation approaches, and is also called Robust Learning Method (RLM) in the following.
	In the evaluation part, the traditional state-of-the-art learning method (SLM) and the RLM are compared, and their robustness is analyzed, if one of the sensor streams is disturbed by unknown noise.


\section{Evaluation}


	In this section, the proposed approaches are evaluated qualitatively on two datasets, the AtUlm-Dataset and the Cityscapes dataset \cite{cityscape_dataset}. 
	The AtUlm dataset is a non public available dataset, which consists of $1446$ fine-annotated grayscales images of a wideangle camera (size $850  \times 1920$), and contains 12 classes (road, sidewalk, pole, traffic light, traffic sign, pedestrian, bicyclist, car, truck, bus, motorcyclist, and background), which are defined in accordance to the Cityscapes dataset.
	The data consists of urban and rural scenarios and was recorded in good weather conditions by our test-vehicle \cite{Kunz_2015_AutonomousDrivingAtUlmUniversityAModularRobustAndSensorIndependentFusionApproach} in the surroundings of Ulm (Germany). Furthermore, for each camera image, the corresponding lidar data of four Velodyne 16  are available, which are mounted on four different positions in the car roof (front, left, right, and rear).
	The dataset is split into two subsets: $1054$ images are used for training, and the remaining $392$ images are used for evaluation. Note, that the training and validation set consists of different sequences recorded at different locations.


	For reproducibility, the different approaches are also evaluated on the Cityscapes dataset \cite{cityscape_dataset}, which consists of several RGB images of 50 German cities and the corresponding depth images and semantic labels.
	Due to the lack of lidar data, the depth image of the stereo camera is used instead of the depth image determined form the lidar data, assuming that camera and stereo data are delivered by two separate sensors and are disturbed independently from each other (knowing that this does not correspond to reality).    
	Similar to other works \cite{Pfeuffer_2019_SemanticSegmentationOfVideoSequencesWithConvolutionalLSTMs, Zhao_2017_ICNet_forRealTimeSemanticSegmentationOnHighResolutionImages, Zhao_2017_PyramidScenParsingNetwork}, 19 of the available 30 classes are used for training. The proposed approaches are trained on the $2975$ fine-annotated camera images and evaluated by means of the $500$ validation images, since the ground-truth of the test-set is not publicly available. 
	
	
	For a fair comparison of the proposed approaches, the training hyper-parameters are chosen identical for each training.
	Each approach is implemented in Tensorflow \cite{tensorflow} and is trained on a single Nivida Titan X on the training set by means of the Stochastic Gradient Descent (SGD) with a momentum equal to $0.9$ and a weight decay of $0.0001$. A batch size of two is chosen due to memory reasons and the initial learning rate is set to $0.001$, which is decreased by means of the poly-learning rate policy. The training loss is identical to the original ICNet implementation \cite{Zhao_2017_ICNet_forRealTimeSemanticSegmentationOnHighResolutionImages}, which is the weighted sum of the cross-entropy losses of each resolution branch.	
	The models are trained for $60k$ iterations in case of the AtUlm dataset and for $100k$ iterations in case of the Cityscapes dataset.
	During training, the images are randomly flipped and randomly resized between $0.5$ and $2.0$ of the origin image size.
	For compensating the training fluctuations, each approach is trained for three times and the average is taken. The training weights of the network are initialized with a pretrained ImageNet network. However, the kernel weights of the first convolutional layer are initialized randomly using a Gaussian distribution, since the input dimension of the Early and Late Fusion varies. 
	Note, that our ICNet implementation differs from the origin implementation described in \cite{Zhao_2017_ICNet_forRealTimeSemanticSegmentationOnHighResolutionImages}, since we do not apply the model compression afterwards, but train the model with half of its feature size.
	Due to this and due to the small batch size, our ICNet implementation achieves a mIoU of $64.7\%$ (batch size 2) on the Cityscapes dataset, while the mIoU of the origin ICNet implementation proposed in \cite{Zhao_2017_ICNet_forRealTimeSemanticSegmentationOnHighResolutionImages} yields $67.7\%$ (batch size 16).
	
	In the next sections, the proposed methods are evaluated by means of pixel-wise accuracy (acc.) and mean Intersection of Union (mIoU). For this, two cases are considered.
	First, the fusion-based approaches, a pure camera-based approach (ICNet\_img) and a pure depth-image-based approach \mbox{(ICNet\_depth)} are compared in good-weather conditions and in scenarios, in which the sensors are disturbed by known noise. Moreover, the robustness of the described methods is tested by means of noise and data artifacts, the neural network has not seen during the training process. 
	Hereinafter, the Late Fusion approach trained by SLM is called LateFusion\_SLM and the Late Fusion approach trained by RLM is denoted as LateFusion\_RLM.

\subsection{Evaluation at Good Weather Conditions}

	The various, proposed approaches are now evaluated in good weather conditions on the AtUlm and Cityscapes dataset using SLM and RLM for training.
	Furthermore, they are also evaluated on a simulated adverse weather dataset, the so-called adverse AtUlm dataset and adverse Cityscapes dataset, to analyses, how the learned models behave towards known disturbances, such as white polygons randomly fitted in either the camera or the depth image. Adverse AtUlm and adverse Cityscapes consist of good weather data of the corresponding dataset ($\approx 14.3\%$), partially disturbed data in which white polygons are randomly fitted in either camera or depth image ($\approx 57.1\%$), and complete impaired data where one of the sensor streams is set to infinity ($\approx 28.6\%$). Each approach is trained on AtUlm, adverse AtUlm, Cityscapes and adverse Cityscapes by means of SLM and RLM. The corresponding results are presented in Table \ref{table_dense_adverseWeather} and Table \ref{table_cityscapes_adverseWeather}.
	Generally, the performance increases the later the sensor data are fused, as results in good and bad weather conditions show. The reason for this is that the sensor data influences each other in an early fusion. If one sensor is disrupted, the complete feature map is disturbed, and hence, the performance of the segmentation approach declines. The Late Fusion approaches deliver the best results for each dataset while ICNet\_depth performs worst. This is caused by the fact that the depth image based approaches can detect the objects at close range, but they cannot classify them correctly. For instance, buses and trucks look similar in the depth data, so that the network cannot distinguish between these two classes. Furthermore, the depth image does only deliver little far-range information, and hence, the correct class cannot be predicted correctly. 
	
	\begin{figure}[tp]
		\includegraphics[width=1.0\columnwidth]{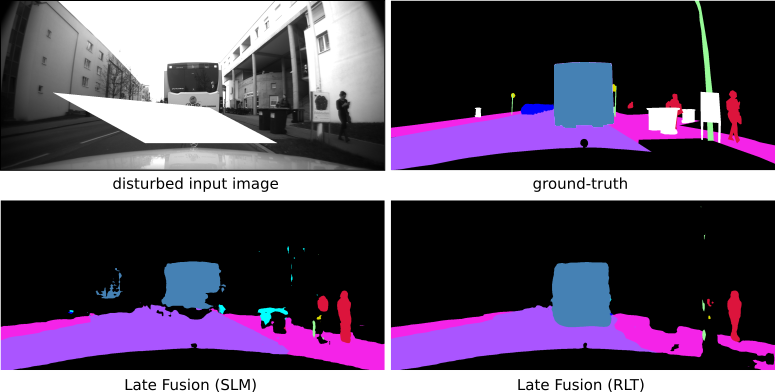}
		\caption{Qualitative comparison of the state-of-the-art training method (SLM) with proposed learning technique (RLM) by means of the adverse AtUlm Dataset.}
		\label{fig_example_dense_lateFusion}
	\end{figure}
	
	The comparison of both learning methods indicates that the RLM-based approaches outperforms the SLM on the adverse datasets, however at the expense of accuracy and mIoU in optimal weather conditions. The reason is that the RLM-based approaches have learned to detect coarse structures in the image, especially at disturbed sensor data, and small image details often get lost, such as the traffic signs in the background of Fig. \ref{fig_example_dense_lateFusion}. Hence, the RLM-based approaches have a lower class-wise accuracy for small objects than the SLM-based methods in good weather conditions (traffic sign:  $56.81\%$ (RLM) versus $67.88\%$ (SLM); traffic light: $56.51\%$ (RLM) versus $58.52\%$ (SLM)), while it is similar or even greater for larger structures (road:  $97.66\%$ (RLM) versus $97.55\%$ (SLM); trucks: $82.96\%$ (RLM) versus $76.95\%$ (SLM)).  
	Nevertheless, the RLM-based approaches are more robust against sensor disturbances and achieve much better results in this case. Additionally, they are less vulnerable to unknown noise as it will be shown in the next section.

	\begin{table}[t]
		\caption{Evaluation on AtUlm Dataset}
		\vspace{-3mm}
		\begin{center}
			\begin{tabular}{|c||c|c||c|c|}
				\hline
				& \multicolumn{2}{|c||}{adverse AtUlm} & \multicolumn{2}{|c|}{AtUlm}  \Tstrut \Bstrut \\ 
				Approach & acc. (\%) & mIoU (\%) & acc. (\%) & mIoU (\%) \Tstrut \Bstrut \\ \hline \Tstrut
				ICNet (img only) & -- & -- & $96.09$ & $51.66$ \Mstrut \\
				ICNet (lidar only) & -- & -- & $92.50$ & $32.02$ \Mstrut \\
				Early Fusion (SLM) & $93.41$  & $42.54$ & $96.17$ & $51.43$ \Mstrut \\
				Early Fusion (RLM) & $94.81$ & $45.27$ & $96.08$ & $50.30$ \Mstrut \\
				Late Fusion (SLM) & $93.66$ & $44.30$ & $\mathbf{96.25}$ & $\mathbf{54.15}$ \Mstrut \\
				Late Fusion (RLM)& $\mathbf{95.28}$ & $\mathbf{46.72}$ & $96.20$ & $52.04$ \Bstrut \\ \hline
			\end{tabular}
			\label{table_dense_adverseWeather}
		\end{center}
		\vspace{-3mm}
	\end{table}

	\begin{table}[t]
		\caption{Evaluation on Cityscapes Dataset}
		\vspace{-3mm}
		\begin{center}
			\begin{tabular}{|c||c|c||c|c|}
				\hline
				& \multicolumn{2}{|c||}{adverse Cityscapes} & \multicolumn{2}{|c|}{Cityscapes}  \Tstrut \Bstrut \\ 
				Approach & acc. (\%) & mIoU (\%) & acc. (\%) & mIoU (\%) \Tstrut \Bstrut \\ \hline \Tstrut
				ICNet (img only) & -- & -- & $93.25$ & $64.73$ \Mstrut \\
				ICNet (depth only) & -- & -- & $86.65$ & $43.65$ \Mstrut \\
				Early Fusion (SLM) & $85.19$ & $51.71$ & $93.26$ & $64.61$ \Mstrut \\
				Early Fusion (RLM) & $90.97$ & $57.07$ & $92.86$ & $63.03$ \Mstrut \\
				Late Fusion (SLM) & $84.80$ & $52.51$ & $\mathbf{93.37}$ & $\mathbf{65.09}$ \Mstrut \\
				Late Fusion (RLM) & $\mathbf{91.57}$ & $\mathbf{58.48}$ & $93.17$ & $64.55$ \Bstrut \\ \hline
			\end{tabular}
			\label{table_cityscapes_adverseWeather}
		\end{center}
		\vspace{-3mm}
	\end{table}

\subsection{Evaluation Using Unknown Disturbances}

	\begin{figure}[tp]
		\includegraphics[width=1.0\columnwidth]{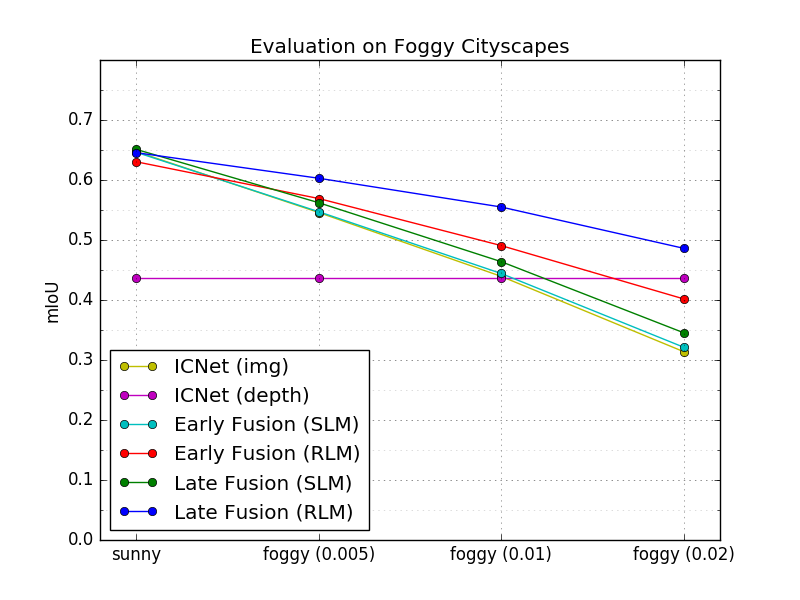}
		\caption{Performance course (mIoU) as function of $\beta$ evaluated on the Foggy Cityscapes dataset}
		\label{fig_results_cityscape_foggy}
	\end{figure}
	
	\begin{table}[t]
		\caption{Evaluation on Foggy Cityscapes}
		\begin{center}
			\begin{tabular}{|c||c|c||c|c|}
				\hline
				& \multicolumn{2}{|c||}{without fog} & \multicolumn{2}{|c|}{dense fog ($\beta=0.02$)}  \Tstrut \Bstrut \\ 
				Approach & acc. (\%) & mIoU (\%) & acc. (\%) & mIoU (\%) \Tstrut \Bstrut \\ \hline \Tstrut
				ICNet (img) & $93.25$ & $64.73$ & $74.09$ & $31.40$ \Mstrut \\
				ICNet (depth) & $86.65$ & $43.62$ & $\mathbf{86.65}$ & $43.62$ \Mstrut \\
				Early Fusion (SLM) & $93.26$ & $64.61$ & $73.25$ & $32.14$ \Mstrut \\
				Early Fusion (RLM) & $92.86$ & $63.03$ & $79.76$ & $40.19$ \Mstrut \\
				Late Fusion (SLM) & $\mathbf{93.37}$ & $\mathbf{65.09}$ & $74.46$ & $34.55$ \Mstrut \\
				Late Fusion (RLM)& $93.17$ & $64.55$ & $86.51$ & $\mathbf{48.62}$  \Bstrut \\ \hline
			\end{tabular}
			\label{table_foggyCityscapes}
		\end{center}
	\end{table}
	
	\begin{figure}[tp]
		\includegraphics[width=1.0\columnwidth]{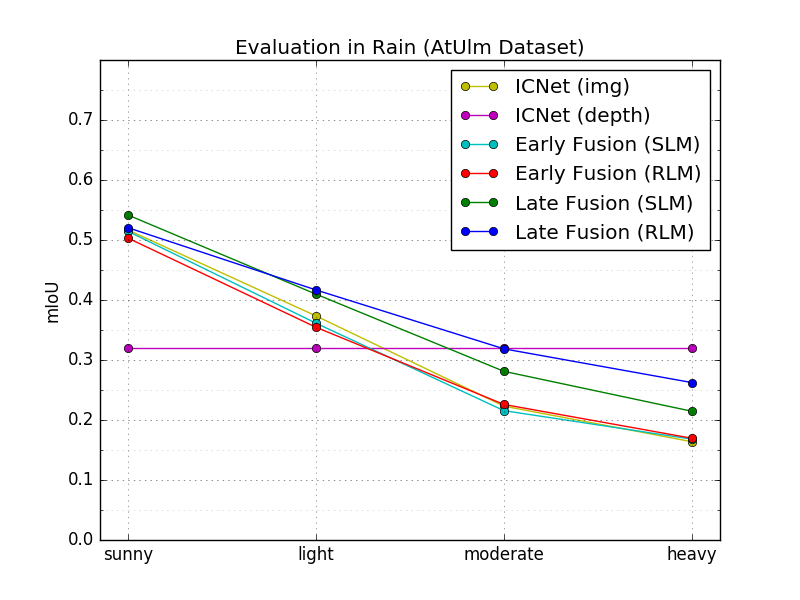}
		\caption{Proposed approaches evaluated at different rain intensities (light, moderate, and heavy rain)}
		\label{fig_results_dense_rain}
	\end{figure}
	
	\begin{table}[t]
		\caption{Evaluation in Rain (AtUlm Dataset)}
		\begin{center}
			\begin{tabular}{|c||c|c||c|c|}
				\hline
				& \multicolumn{2}{|c||}{sunny} & \multicolumn{2}{|c|}{heavy rain}  \Tstrut \Bstrut \\ 
				Approach & acc. (\%) & mIoU (\%) & acc. (\%) & mIoU (\%) \Tstrut \Bstrut \\ \hline \Tstrut
				ICNet (img) & $96.09$ & $51.66$ & $84.77$ & $16.38$ \Mstrut \\
				ICNet (depth) & $92.50$ & $32.02$ & $\mathbf{92.50}$ & $\mathbf{32.02}$ \Mstrut \\
				Early Fusion (SLM) & $96.17$ & $51.43$ & $83.76$ & $16.81$ \Mstrut \\
				Early Fusion (RLM) & $96.08$ & $50.30$ & $85.64$ & $16.96$ \Mstrut \\
				Late Fusion (SLM) & $\mathbf{96.25}$ & $\mathbf{54.15}$ & $88.31$ & $21.46$ \Mstrut \\
				Late Fusion (RLM)& $96.20$ & $52.04$ & $91.17$ & $26.24$  \Bstrut \\ \hline
			\end{tabular}
			\label{table_Rainy_Dense}
		\end{center}
	\end{table}

\begin{figure*}[tbp]
		\includegraphics[width=1.0\textwidth]{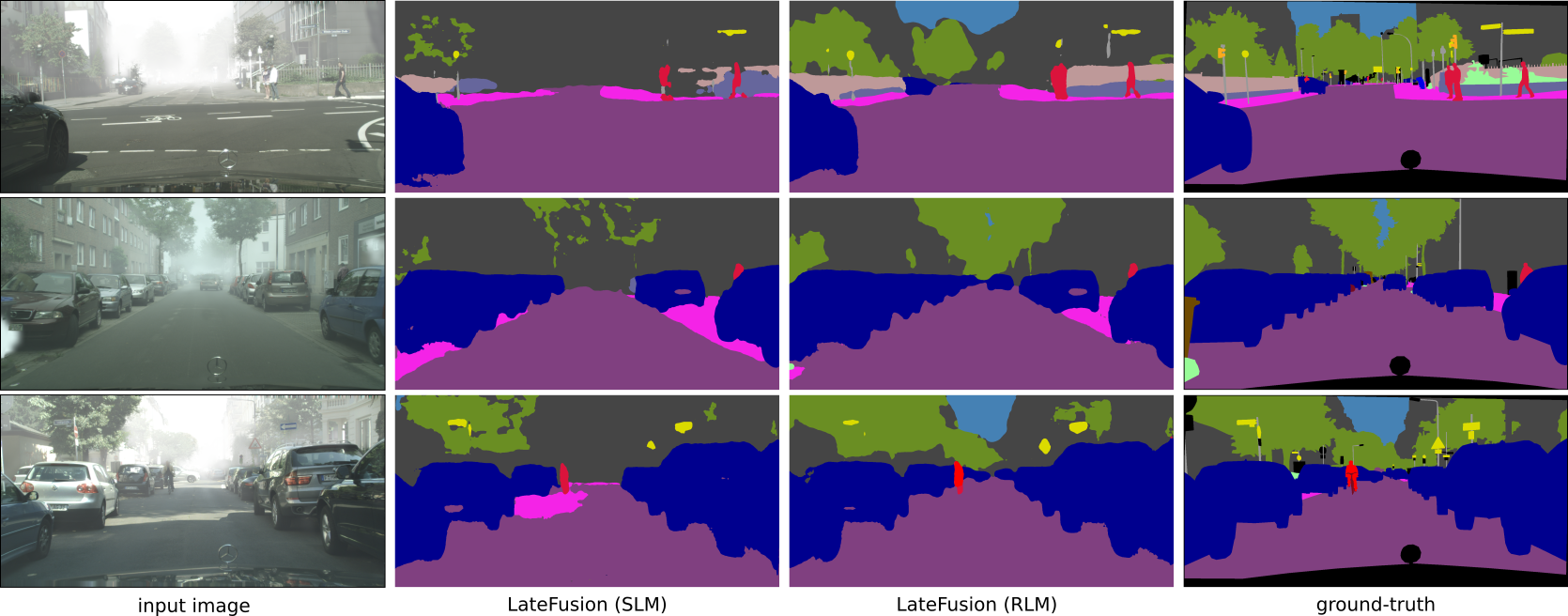}
		\caption{Qualitative results on the Foggy Cityscapes dataset ($\beta=0.020$). First column: input image; second column: 
		results of the Late Fusion approach using the standard training methods (SLM); third column: results of the Late Fusion approach using the Robust Learning Method (RLM); fourth column: corresponding ground-truth.}
		\label{fig_Foggy_Cityscapes_examples}
	\end{figure*}

	In this section, the proposed fusion approaches are evaluated how they behave if an unknown disturbance occurs in the sensor data and three different noise types (fog, blinding sun, and rain) are considered in more detail. First, the trained models are evaluated in fog by means of the Foggy Cityscapes dataset \cite{Sakaridis_2018_SemanticFoggySceneUnderstandingWithSyntheticData}.
	Foggy Cityscapes is a synthetic fog dataset, which is derived from the Cityscapes Dataset. The real-world images of the Cityscapes dataset are modified by adding synthetic fog of varying density. The fog density and the corresponding visibility range depends on a constant attenuation coefficient $\beta$. In the dataset, foggy images with $\beta \in \{0.005; 0.010; 0.020\}$ are provided, which correspond to visibility ranges of $800m$, $400m$ and $200m$ respectively.
	The proposed fusion approaches, ICNet\_img and ICNet\_depth are trained on the Cityscapes dataset by means of SLM and RLM, and are evaluated on the Foggy Cityscapes for each $\beta$. 
	In Fig. \ref{fig_results_cityscape_foggy}, their mIoU-values are shown as a function of $\beta$.
	Their performances decrease the denser the fog is except for ICNet\_depth, whose performance is constant, since it only depends on the depth image, and hence is not disturbed by the fog in the camera image. 
	However, the performances decrease differently. For approaches trained by SLM the performance decreases much more, and at dense fog ($\beta=0.020$), their mIoU-values are even lower than the pure depth based approach. In contrast, the RLM-trained methods perform much better, e.g. LateFusion\_RLM outperforms the other approaches by at least $15\%$ and ICNet\_depth by about $5\%$ (see Table \ref{table_foggyCityscapes} for more details).  
	Furthermore, the proposed methods are also evaluated qualitatively on Foggy Cityscapes. Fig. \ref{fig_result_introduction} and Fig. \ref{fig_Foggy_Cityscapes_examples} show some examples of the Late Fusion approach trained with SLM and RLM using the densest fog ($\beta = 0.020$). It turns out that both learning techniques perform similarly at close range, while RLM outperforms the other one in the long-distance range. For instance, the RLM still detects vehicles slightly covered by fog. In contrast, both approaches fail to predict small objects such as traffic signs occurring in the distance.

\begin{figure*}[tp]
		\includegraphics[width=1.0\textwidth]{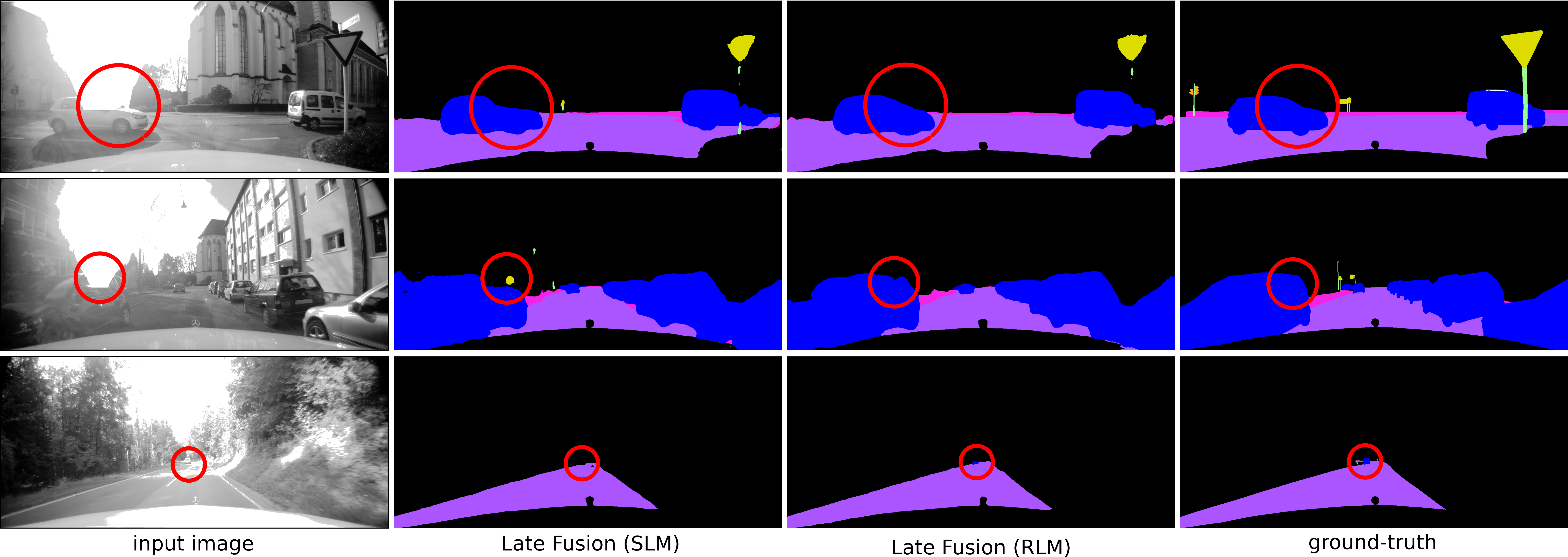}
		\caption{Qualitative results on the on the AtUlm dataset (blinding sun). First column: input image; second column: results of the Late Fusion approach using the standard training method (SLM); third column: results of the Late Fusion approach using the Robust Learning Method (RLM); fourth column: corresponding ground-truth}
		\label{fig_Dense_examples}
	\end{figure*}

\begin{figure*}[tp]
		\includegraphics[width=1.0\textwidth]{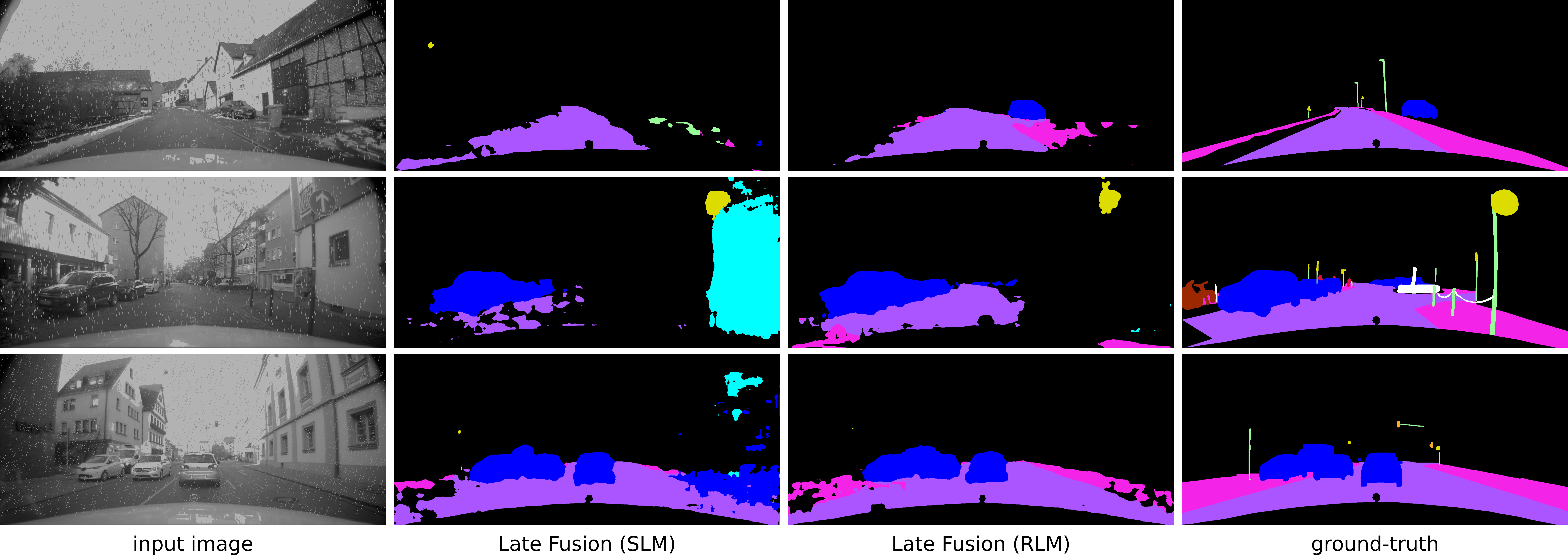}
		\caption{Qualitative results on the on the AtUlm dataset (rain). First column: input image; second column: results of the Late Fusion approach using the standard training method (SLM); third column: results of the Late Fusion approach using the Robust Learning Method (RLM); fourth column: corresponding ground-truth}
		\label{fig_Dense_examples_rain}
	\end{figure*}

	The RLM-based approaches are also more robust, if the camera is blinded by the sun, as it will be shown on sunny images of the AtUlm dataset. In case of dazzling sun, parts of the image are disturbed by white spots. The state-of-the-art methods do not assign the disturbed pixels as car, while the RLM-based methods predict most disturbed pixels correctly, as shown in Fig \ref{fig_Dense_examples}. 
	Nevertheless, the RLM-based approaches also have problems to detect vehicles, which are far away, such as the car in the last row of Fig \ref{fig_Dense_examples}. However, this is mainly caused by sensor limitations, since the used lidars \mbox{(Velodyne 16)} deliver only one or two points for these vehicles, and hence, a precise classification is not possible any more. A more accurate lidar will increase the classification rate in far range further. 
	Due to the lack of real blinding sun data, a qualitative evaluation is not possible. Instead, a second use case is considered, in which synthetic rain is added to the real-world images of the AtUlm dataset. The rain is simulated by randomly drawing $N$ small lines of pixel length $l$ in the camera image, which randomly chosen slant remains constant in one image. Furthermore, the image brightness is reduced by $30\%$, since rainy days are usually darker. 
	For evaluation, different rain intensities are simulated, namely light ($N=500$, $l=10$), moderate ($N=1500$, $l=30$), and heavy ($N=2500$, $l=60$) rain. In Fig. \ref{fig_results_dense_rain}, the mIoU course of the considered approaches is shown in dependence of the rain intensity, and in Table \ref{table_Rainy_Dense}, the results of heavy rain are shown. The graph shows that the performance of all approaches decreases the heavier the rain is. LateFusion\_RLM again performs best, but in heavy rain it is outperformed by ICNet\_depth. The reason is that ICNet\_depth only uses the lidar data, which is not disturbed by the rain drops in this work. In Fig. \ref{fig_Dense_examples_rain}, LateFusion\_SLM and LateFusion\_RLM are compared qualitatively in the case of heavy rain. Generally, the qualitative analysis shows that the LateFusion\_RLM is more robust to the rain. For instance, the RLM-based approach detects the cars more reliably (see first and second row of Fig. \ref{fig_Dense_examples_rain}) and there are much less areas, which are misclassified (e.g. see second and third row of \mbox{Fig. \ref{fig_Dense_examples_rain}}). However, small details such as poles or bicycles are not detected in heavy rain any more, which is mainly caused that these objects are difficult to observe.

\section{Conclusion}

	In this paper, both performance and robustness of semantic segmentation approaches were investigated in adverse weather conditions. It turns out that they decrease enormously in this case as compared to good weather conditions. This problem can be solved in two steps. First, the segmentation accuracy can be increased by fusing different sensor data, e.g camera and lidar data, and so two different sensor data fusion architectures were presented. It showed that the Late Fusion approach outperforms the Early Fusion approach, especially in diverse weather conditions. However, these approaches do not deliver reliable results in diverse weather conditions, since they are vulnerable to unknown noise. 
	Hence, in a second step, the proposed network architectures were trained by means of the Robust Learning Method, which makes the approaches more robust in adverse weather conditions. More concretely,  
	both robustness and performance could be increased enormously by this learning method, as demonstrated on several examples. Furthermore, it was shown, that the proposed approaches also perform well, if unknown disturbances occur that were not presented during training.

\section{Acknowledgment}
	
	The research leading to these results has received funding from the European Union under the H2020 ECSEL Programme as part of the DENSE project, contract number 692449.

\bibliographystyle{plain}
\bibliography{/home/andreas/Documents/Literatur/Jabref-Datebase/Literatur_Promotion}

\end{document}